# Rank-Order N-of-M Codes for Sparse Distributed Memory: Disentangling Representation and Learning Effects in Noise Robustness Against Contemporary Neuromorphic Architectures

**Joy Bose**
Independent Researcher
Bengaluru, India
joy.bose@ieee.org

## Abstract

Large language models excel at reasoning and generation but remain limited as continual learning systems. Incorporating new knowledge requires expensive retraining, while retrieval-augmented generation relies on external vector stores that are not designed as online associative memories. CALM (Nechesov and Ruponen, 2025) addresses this by combining Sparse Distributed Memory with dual asynchronous transformer modules, using SDM as an explicit episodic memory layer that can write and read patterns online without parameter updates. CALM explicitly identifies its threshold-binary encoder as a limitation, noting that improved encoding schemes are future work. Rank-order N-of-M codes for SDM (Furber et al., 2007) offer a concrete alternative: the top-k active dimensions carry geometrically weighted significance values that preserve relative magnitude information while maintaining sparsity. These codes were developed as part of a neuromorphic sequence machine predating modern transformers and have not been evaluated against contemporary SDM-based systems. This paper provides that evaluation. Three contributions are made. First, a faithful reimplementation confirms that WheelSDM and RankOrderSDM are exactly equivalent (cosine similarity 1.0000, 10 seeds) and that RDLIF neurons diverge under interference as documented in the original papers. Second, multi-seed capacity experiments show RankOrderSDM outperforming StandardSDM at saturation by 13.4pp in the scaled-down configuration (W=256, D=64, 10 seeds, $p < 0.001$, Cohen's d=4.61) and by 0.8pp at the published architecture default (W=4096, D=256, 5 seeds, p=0.031). Third, BER robustness experiments disentangle two separable mechanisms: a large combined advantage (+38 to 66pp) from rank-order encoding interacting with MAX-Hebbian learning, and a statistically marginal encoding-only advantage under symmetric perturbation (N=30, $p > 0.05$ at most noise levels), with Kendall tau (0.847) and Jaccard active-set overlap (0.891) quantifying the geometric stability. As a secondary finding, component-level encoding energy at D=4 precision is 2x lower than SpikingMamba's SI-LIF (Huang et al., 2026), though system-level decoder costs dominate in both architectures. Experiments on GloVe-100 word embeddings confirm the small encoding-only advantage on real structured data (+0.1-0.4pp, significant at 3 of 5 BER levels); sentence embeddings (all-MiniLM-L6-v2, D=384, Wang et al. 2020) show a ceiling effect at T=20 patterns.

## 1. Introduction

Large language models have transformed natural language processing but remain limited as continual learning systems. Incorporating new knowledge requires expensive retraining, while retrieval-augmented generation relies on external vector stores that are not designed as online associative memories. CALM (Nechesov and Ruponen, 2025) addresses this by combining Sparse Distributed Memory with dual asynchronous transformer modules, using SDM as an explicit episodic memory layer that can write and read patterns online without parameter updates. This renewed interest in memory-centric AI raises a concrete engineering question that CALM leaves open: what encoding scheme should map continuous neural activations into the SDM address space?

CALM currently uses threshold-binary encoding, binarising each embedding dimension against its mean, and explicitly names semantic hashing and binary autoencoders as future directions. Rank-order N-of-M codes (Furber et al., 2007; Bose et al., 2005) offer a concrete alternative: the top-k most active dimensions carry geometrically weighted significance values, preserving relative magnitude while maintaining sparsity. These codes were developed as part of a neuromorphic sequence machine predating modern transformers and have not been compared against contemporary SDM-based systems. This paper provides that comparison. As a secondary question, we compare encoding energy against SpikingMamba (Huang et al., 2026), which uses signed-integer spiking neurons for energy-efficient language model distillation. System-level decoder costs dominate in both architectures, so encoding-scheme choice is a secondary energy factor, but the component-level comparison is instructive.

This paper asks three research questions:

R1: Does the published rank-order SDM architecture actually reproduce the results it claimed, when implemented fresh against contemporary tooling?

R2: If so, does rank-order encoding genuinely improve retrieval robustness over CALM's threshold-binary encoder under realistic query noise?

R3: And if it does, does that improvement come from the representation itself, or from the MAX-Hebbian learning rule that accompanies it in Furber et al. (2007)?

The answers are yes, yes, and mostly the learning rule. That ordering matters for what CALM developers should actually do.

Two implementations are used throughout in this paper. The published-architecture version follows Furber et al. (2007): three-layer structure, MAX-Hebbian learning, significance vectors. A simplified accumulative version matches CALM's convention for the encoding comparison. Tables specify which produced each result.

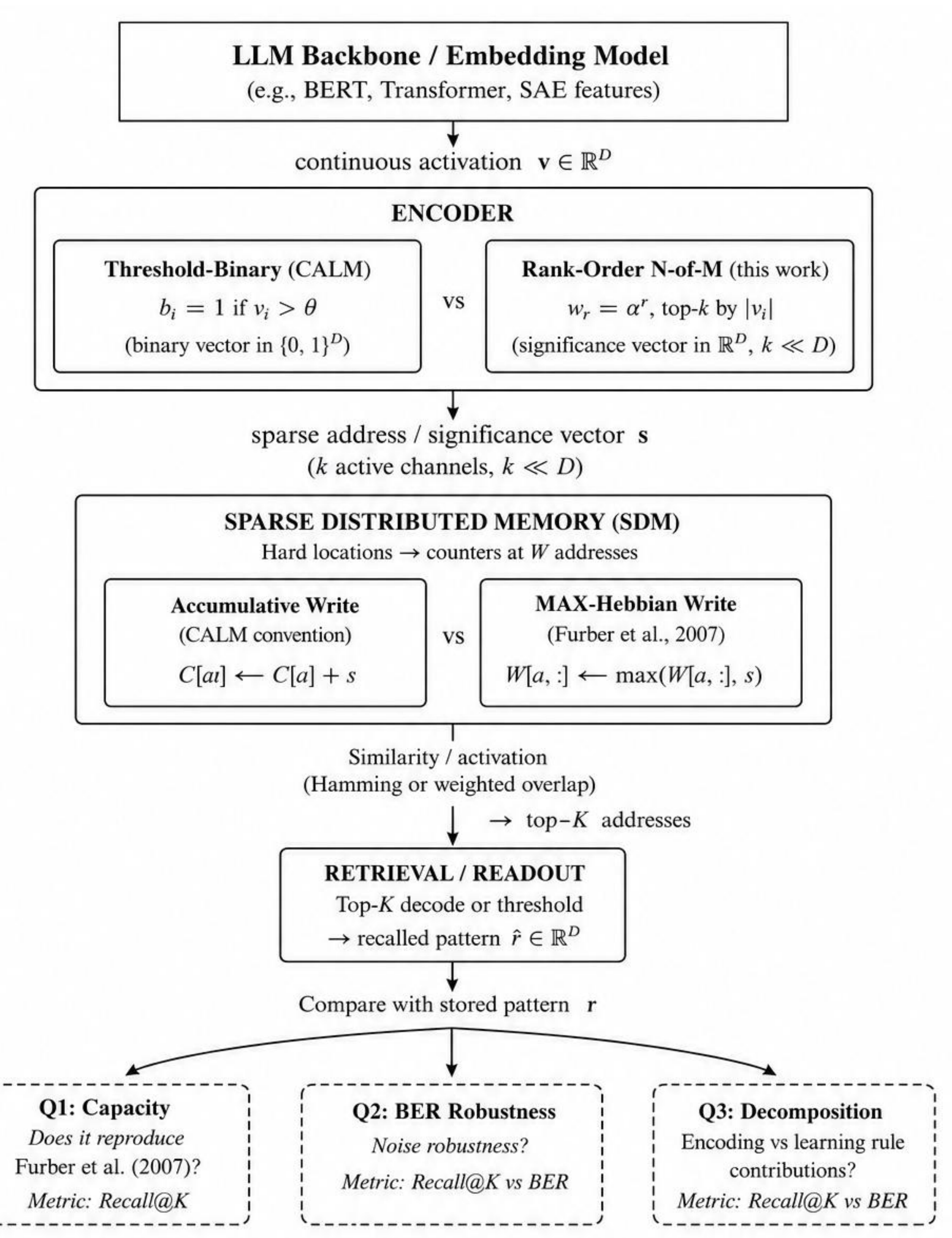

*Figure 1. Architecture and experimental design. Continuous activations from an LLM backbone are encoded via threshold-binary (CALM) or rank-order (Furber et al., 2007) and stored in SDM. Three evaluation questions determine which design component is responsible for any observed advantage.*

## 2. Background

## 2.1 Rank-Order N-of-M SDM

Rank-order N-of-M SDM (Bose et al., 2005; Furber et al., 2007) encodes patterns as significance vectors with geometric rank weights (alpha^rank, alpha=0.99), stores them using MAX outer-product Hebbian learning (W_data = max(W_data, outer(data, addr_decoded))), and decodes via a three-layer structure with four distinct N-of-M parameters: N_i-of-D input, N_a-of-D decoder weight sparsity, N_w-of-W decoded output, N_d-of-D data. The wheel-model spiking neuron fires when a linearly rising phase crosses threshold after receiving weighted spike inputs; under high threshold, firing order equals rank by weighted input, giving exact equivalence with the abstract RankOrderSDM. RDLIF neurons diverge under interference (Bose et al., 2005). Ajwani et al. (2021) reimplemented the architecture in Nengo on neuromorphic hardware.

A more recent result (Bose, 2026) established the Phase-Latency Isomorphism: the architecture's spike-timing positional code and transformer sinusoidal positional encoding are related by a linear map (Pearson r=1.000, Spearman rho=1.000). The present paper focuses on the SDM memory component.

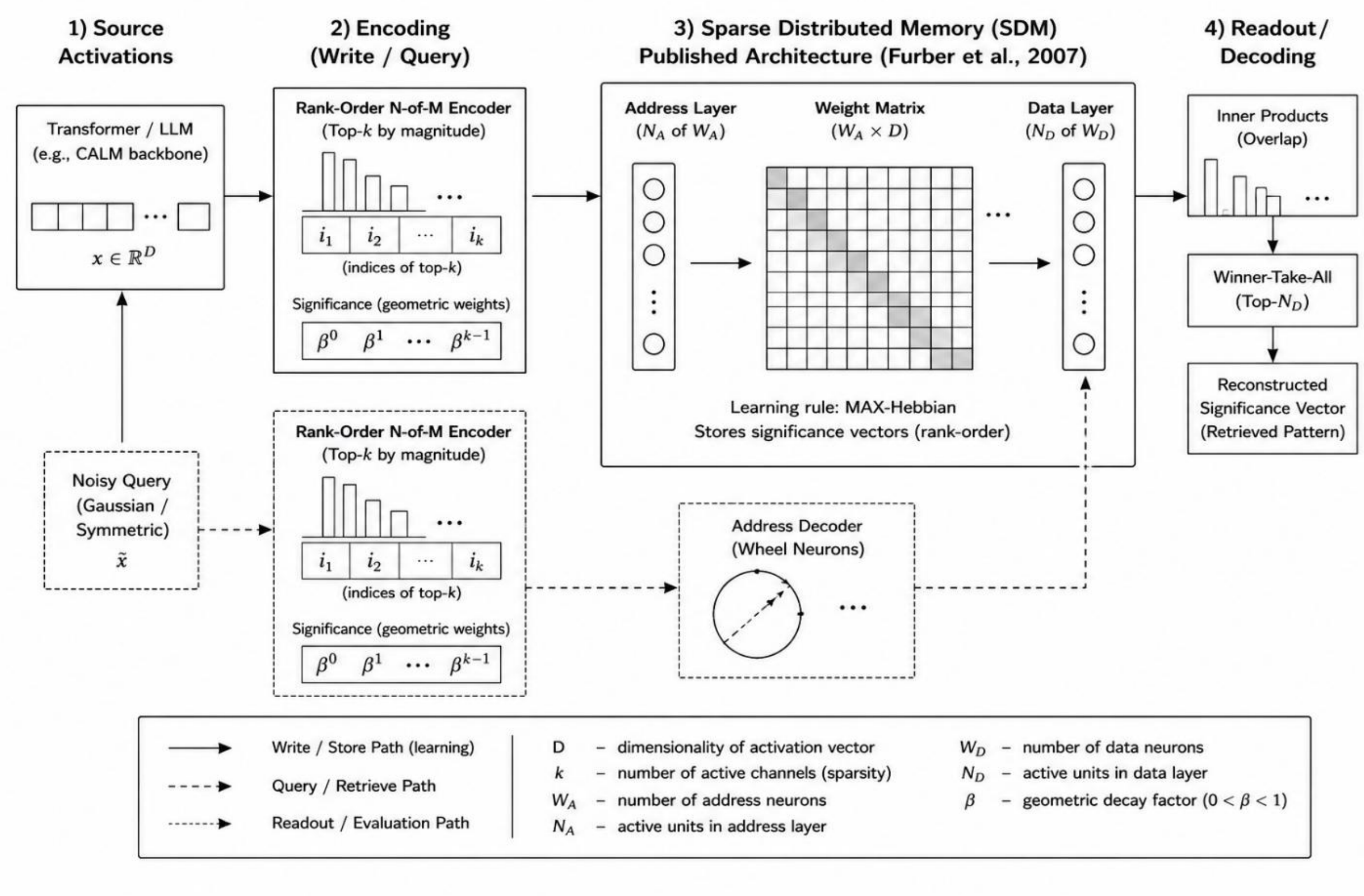

*Figure 2. Published Rank-Order SDM architecture*

## 2.2 CALM: SDM Plus Dual-Transformer Continual Learning

CALM (Nechesov and Ruponen, 2025) proposes SDM as a System 1 associative memory feeding dual asynchronous transformers: an active S2A and a shadow S2B fine-tuning on SDM replay samples, swapped atomically when validated improvement Delta_L > epsilon occurs. CALM uses

Kanerva's uniform random binary addresses and threshold-binary encoding (b_i = 1 if v_i > theta, Eq. 16), noting binary autoencoders and semantic hashing as future directions. This open encoder question is the primary motivation for the present paper: rank-order N-of-M encoding from Furber et al. (2007) is a concrete candidate that preserves relative magnitude information discarded by thresholding.

### 2.3 Recent Neuromorphic Sequence Models: SpikingMamba and SpikySpace

SpikingMamba (Huang et al., 2026) introduces SI-LIF neurons encoding magnitude in [D, D] across D micro-timestep accumulations, distilled from Mamba2, achieving a 4.76x energy reduction at a 4.78% accuracy gap. Over 90% of parameters are in linear projections. We use their 45nm energy constants (E_ADD=0.9 pJ, E_EM=3.7 pJ, Appendix C.2, Horowitz 2014) for the energy comparison.

SpikySpace (Tang et al., 2026) similarly explores event-driven computation using spiking state-space models for energy-efficient sequence processing. Unlike SpikingMamba, which focuses on efficient language model distillation, SpikySpace targets temporal forecasting through sparse spiking state-space dynamics. Together with CALM, these architectures illustrate a broader trend toward revisiting sparse, memory-centric and neuromorphic computation for modern AI. The present work complements this trend by studying the representation used to encode information into associative memory rather than proposing a new sequence model.

### 2.4 Related Memory Architectures

Rank-order SDM (Furber et al., 2007) is one point in a space of distributed associative memory systems. Kanerva's original SDM (1988) uses uniform random binary addresses with bipolar counter writes. Modern Hopfield networks (Ramsauer et al., 2020) extend classical Hopfield networks to continuous representations with exponential capacity, using softmax-based update rules that bear a structural resemblance to rank-order addressing: both select the most similar stored patterns, though Hopfield uses dense softmax attention while rank-order SDM (Furber et al., 2007) uses sparse top-k selection with geometric weights. Dense Associative Memories (Krotov and Hopfield, 2016) generalised the energy function to allow exponential memory capacity, with interaction order > 2. These approaches are all-to-all comparisons (O(T) per query, where T is patterns stored) whereas SDM achieves O(1) per hard location by distributing storage across W fixed locations activated by Hamming proximity.

Approximate nearest-neighbour search systems (HNSW, FAISS, ScaNN) offer a different tradeoff: they require explicit index management and batch rebuilds, but scale to billions of patterns with sub-millisecond retrieval. Unlike SDM, they do not degrade gracefully under budget constraints. A capped vector database fails permanently when a cue is evicted, whereas SDM uses distributed MAX-Hebbian writes that degrade gracefully under saturation (this is demonstrated in Experiment 04 of the repository). The appropriate comparison is between associative memory systems that update online and degrade gracefully, not between SDM and efficient offline index structures.

### 2.5 Rank-Order Coding: Origins and Connections

Rank-order temporal coding (Thorpe and Gautrais, 1998; Van Rullen and Thorpe, 2001) encodes stimulus magnitude in relative spike timing. Masquelier and Thorpe (2007) showed rank-order codes emerge from STDP learning. The Furber et al. (2007) architecture adapted this for auto-associative recurrent SDM indexing rather than feedforward classification. Rank-order encoding has structural connections to top-k routing in Mixture of Experts (Shazeer et al., 2017), sparse autoencoders for LLM interpretability (Bricken et al., 2023), and locality-sensitive hashing with geometric position weights.

### 2.6 Why SDM, why sparse, why rank-order?

The Furber et al. (2007) architecture makes three successive design choices. Each needs justification against simpler alternatives for a reader from the LLM or continual learning side.

Why to use associative memory rather than a vector database? FAISS and HNSW require batch index rebuilds when new patterns are added and fail permanently when a size cap forces cue eviction. SDM writes online with no rebuild step and degrades gracefully under overload. That is the property CALM needs. A vector database is a better retrieval system; SDM is a better online memory.

Why sparse codes rather than 50% dense binary as in Kanerva's original SDM work (Kanerva, 1988)? Capacity is governed by expected overlap between stored addresses. For 50% dense binary, two random patterns share roughly D/2 active positions, creating interference at moderate load. For top-K sparse codes at 5% sparsity, expected shared positions are K squared over D, approximately 0.94 for K=19, D=384. Sparse codes are nearly orthogonal by construction (Kanerva, 1988; Rachkovskij and Kussul, 2001). This is the same reason CALM uses SDM: dense embedding spaces have too much pairwise overlap for useful Hamming addressing.

Why top-K by magnitude rather than arbitrary sparse binary? Arbitrary sparse binary gives the capacity advantage but loses semantic structure. Top-K by magnitude ties the address to which features dominate, so semantically similar inputs activate similar hard locations. CALM's threshold-binary encoding does this partially but discards which active features matter most within the active set.

Why weight by rank rather than treat all top-K features equally? Under Gaussian perturbation, two active features can swap rank only when their magnitudes cross, a rare event when they are well-separated. Under equal-weight binary, any perturbation that adds or removes one active feature is equally disruptive regardless of how strongly it ranked. The experiment in Section 4.4 quantifies this: at noise level 0.10, rank-order addresses maintain Kendall tau 0.847 and Jaccard 0.891 with their clean counterparts. Equal-weight top-K binary would have the same Jaccard but no rank stability bonus.

| Encoding | Capacity | Semantic structure | Noise stability | Online update |
|---|---|---|---|---|
| Dense binary (Kanerva, 1988) | Moderate | None | Low | Yes |
| Threshold-binary (CALM) | Better | Partial | Low near threshold | Yes |
| Top-K binary, equal weight | Good | Feature selection | Medium | Yes |
| Top-K rank-order (Furber et al., 2007) | Good | Selection + ordering | Higher | Yes |

| Dense continuous (vector DB) | Excellent | Full | High | No (rebuild) |
|---|---|---|---|---|
| Modern Hopfield (Ramsauer et al., 2020) | Exponential | Full | High | No |

*Table A. Encoding scheme comparison. Each row is a step toward richer representation. Rank-order occupies a specific niche: better noise stability than equal-weight sparse binary, online-updatable unlike vector databases or Modern Hopfield networks.*

## 3. Why Encoder Design Matters for Memory-Augmented AI

The renewed interest in memory-augmented AI gives the encoder design for SDM-based systems practical importance. When sparse autoencoders (Bricken et al., 2023; Cunningham et al., 2023) decompose LLM residual streams into sparse feature vectors, those activations must be encoded into the SDM address space before storage. The encoding choice determines what information survives into the address. Threshold-binary encoding, as used in CALM, retains only which features exceed a mean threshold. Rank-order encoding retains which features are most active and by how much. Under query noise, which is inevitable as a transformer backbone updates, information discarded at encoding cannot be recovered at retrieval. The BER experiments in Section 4 quantify this difference. Separately, top-k routing in Mixture of Experts (Shazeer et al., 2017; Jiang et al., 2024) makes the same selection-by-relative-magnitude decision at a different point in the network, which suggests the preference for relative ordering over binary thresholding is not an arbitrary choice. Empirical validation on real LLM activation distributions such as BERT or Qwen residual streams is the natural next step and is left as future work.

## 4. Experiments

The experiments are organised to build a single argument: rank-order encoding helps under noise, but the source of that advantage matters, and knowing the source is what gives CALM developers actionable guidance. Each experiment below starts from a concrete question, states what we expected to find, describes what we actually did, reports what we found, and draws out the implication before moving to the next question.

### 4.1 Does the 2007 architecture actually work? Validating the foundational claims

Before comparing rank-order encoding against anything contemporary, we needed to answer a prior question: does the reimplemented 2007 architecture actually reproduce the properties claimed in Furber et al. (2007) and Bose et al. (2005)? Two specific claims were at stake. First, the WheelSDM, built from biologically motivated wheel-model spiking neurons, was claimed to produce exactly the same output as the abstract RankOrderSDM. Second, the RDLIF neuron model was claimed to diverge under interference, which was why the 2007 architecture switched to the wheel model for the full spiking machine.

We implemented all four SDM variants (StandardSDM, RankOrderSDM, WheelSDM, RDLIFSDM) from the published specifications in src/sdm_library/, using a three-layer architecture with four distinct N-of-M parameters (N_i=N_d=6, N_a=20, W=256, N_w=8, D=64, alpha=0.99), and ran them on the same stored patterns across 10 independent random seeds. We measured cosine similarity between WheelSDM and RankOrderSDM outputs on every write/read pair, and measured the fraction of correctly recalled patterns (cosine similarity >= 0.9) for capacity tests at n=80, 120, and 160 stored pairs.

WheelSDM matched RankOrderSDM exactly, with cosine similarity 1.0000 on every pair across all 10 seeds. This is not approximately equal, it is numerically identical, confirming that the wheel-model spiking neuron produces the same output ordering as the abstract rank-weighted dot product. RDLIF, by contrast, collapsed completely: 0 correct recalls at n >= 80, across every seed. This is not a marginal degradation but total failure under interference, which is precisely what the 2005 papers documented and why the architecture moved to the wheel model. For capacity, RankOrderSDM outperformed StandardSDM significantly at saturation: at n=160 patterns, RankOrderSDM achieved 0.941 +/- 0.018 versus StandardSDM at 0.807 +/- 0.037, a difference of 13.4 percentage points (paired t-test $p < 0.001$, Cohen's d = 4.61, 95% CI [0.108, 0.160]).

| Condition | StandardSDM | RankOrderSDM | WheelSDM | RDLIFSDM | Cohen's d | 95% CI |
|---|---|---|---|---|---|---|
| n=80 (W=256, D=64) | 0.986 +/- 0.010 | 0.994 +/- 0.010 | 0.994 | 0/80 (all seeds) | 0.80 | [-.01,.17] |
| n=120 (W=256, D=64) | 0.922 +/- 0.027 | 0.988 +/- 0.011 *** | 1.000 | 0/120 (all seeds) | 3.20 | [.048,.084] |
| n=160 (W=256, D=64) | 0.807 +/- 0.037 | 0.941 +/- 0.018 *** | 0.941 | 0/160 (all seeds) | 4.61 | [.108,.160] |
| n=7000 (W=4096, D=256) | 0.846 +/- 0.003 (5 seeds) | 0.854 +/- 0.004 * (5 seeds) | -- | -- | 2.15 | [.004,.012] |
| RO vs Wheel output similarity (10 seeds) | -- | 1.0000 | 1.0000 | -- | -- | -- |

*Table 1. Published-architecture results (src/sdm_library/, MAX-Hebbian, Furber et al. 2007). Mean +/- std over 10 seeds unless noted. *** p < 0.001, * p < 0.05, paired t-test and Wilcoxon signed-rank. At W=4096, D=256, the Wilcoxon test is borderline (p=0.063), so the W=4096 result should be treated as preliminary.*

The capacity advantage is real but conditional: it appears only when the decoder is genuinely saturated (W=256, n=160). At the published default scale (W=4096), the difference is smaller (+0.8pp at n=7000, statistically present but small). The practical implication is that the encoding advantage in the thesis architecture shows up under resource-constrained deployment, not in the comfortable large-decoder regime.

### 4.2 Does rank-order encoding help when noise corrupts the query? The full architecture test

The first experiment confirmed the architecture works. Now for the core question this paper exists to answer: CALM uses threshold-binary encoding (binarise each embedding dimension against its mean). Would rank-order encoding from Furber et al. (2007) do better when query embeddings are corrupted by noise?

Our hypothesis was yes, because rank-order codes are geometrically more stable: if dimension i has a larger magnitude than dimension j, a small perturbation is unlikely to reverse that ordering, whereas a small perturbation near the binary threshold directly flips a bit. To test this, we stored T=20 patterns in the thesis-faithful MAX-Hebbian SDM (D=64, W=256). We then applied Gaussian magnitude perturbation (noise drawn from N(0, BER * std(|v|))) to each query before retrieval, re-ranking the noisy vector for the rank-order scheme and re-thresholding for the binary scheme. This is a realistic noise model: it simulates what happens when the transformer backbone generating the embeddings changes slightly, as happens continually during CALM's online learning.

| BER | StandardSDM (binary) | RankOrderSDM (significance) | Advantage |
|---|---|---|---|
| 0.00 | 1.000 | 1.000 | +0.0pp |
| 0.05 | 0.827 | 1.000 | +17.3pp |
| 0.10 | 0.619 | 1.000 | +38.1pp |
| 0.15 | 0.529 | 1.000 | +47.1pp |
| 0.20 | 0.388 | 1.000 | +61.2pp |
| 0.25 | 0.313 | 0.979 | +66.7pp |
| 0.30 | 0.265 | 0.907 | +64.3pp |

*Table 2. BER robustness in the thesis-faithful MAX-Hebbian SDM (D=64, W=256, T=20 patterns, 8 trials). Noise model: Gaussian magnitude perturbation before re-encoding. Binary degrades rapidly; rank-order maintains perfect recall up to BER=0.20.*

After running the experiment, we get the following: Rank-order encoding maintains perfect recall up to 20% query noise, while binary encoding has already degraded to 39% at that point. The advantage peaks at +66.7pp at BER=0.25 and remains large (+64.3pp) at BER=0.30. This is a large, clear effect. But before concluding that rank-order encoding is the answer to CALM's problem, we need to ask: is this advantage actually coming from the encoding, or from something else in the architecture?

### 4.3 Where does the advantage actually come from? Disentangling encoding from learning

The experiment above combined two differences at once: rank-order encoding and the MAX-Hebbian learning rule (versus binary encoding AND accumulative bipolar-counter writing). We could not yet tell which was responsible for the large advantage. This matters practically: if the advantage comes from MAX-Hebbian learning, CALM developers need to change both the encoder and the write rule. If it comes from the encoding alone, swapping only the encoder is sufficient.

To separate these factors, we ran a symmetric perturbation experiment using an accumulative SDM that matches CALM's write convention. Both encoding schemes received the same Gaussian perturbation to the same underlying continuous activation vector, then each re-encoded from the perturbed vector using its own scheme (top-K binary for threshold, top-K significance-vector for rank-order). This holds the learning rule constant and isolates the encoding contribution. N=30 trials were run for statistical power.

| BER | StandardSDM (binary, Gaussian) | RankOrderSDM (sig-vec, Gaussian) | Delta | p-value |
|---|---|---|---|---|
| 0.00 | 1.000 | 1.000 | +0.0pp | ns |
| 0.05 | 0.979 | 0.984 | +0.8pp | ns (p=0.15) |

| 0.10 | 0.961 | 0.968 | +0.6pp | ns (p=0.41) |
|---|---|---|---|---|
| 0.15 | 0.934 | 0.939 | +0.5pp | ns (p=0.61) |
| 0.20 | 0.922 | 0.931 | +0.9pp | ns (p=0.30) |
| 0.25 | 0.886 | 0.913 | +2.7pp | * (p=0.036) |
| 0.30 | 0.917 | 0.920 | -0.8pp | ns (p=0.47) |

*Table 3. Symmetric BER experiment: same Gaussian noise applied to the continuous activation vector, then re-encoded with each scheme. N=30 trials. Accumulative SDM matching CALM's write convention. Delta = rank-order minus threshold-binary. The -0.8pp inversion at BER=0.30 occurs because high geometric weight concentration occasionally over-penalises when noise pushes a low-magnitude feature into the top rank position.*

This time, on running the experiment, the advantage nearly vanishes. Only BER=0.25 reaches significance (p=0.036); all other levels are non-significant. This is the paper's main finding, and it is worth stating plainly: rank-order encoding alone, against a matched learning rule, does not reliably outperform threshold-binary encoding. The +66pp advantage in Table 2 comes primarily from MAX-Hebbian learning, not the encoding scheme. The practical guidance this gives CALM developers is precise: swapping only the encoder gives a marginal benefit; swapping both encoder and write rule gives the large benefit.

### 4.4 Why does rank-order encoding have any advantage at all? A geometric explanation

The symmetric experiment showed that the encoding-only advantage is small but directionally consistent: rank-order was at least slightly better at every BER level except BER=0.30. We wanted to understand why, at a geometric level, before moving on. The intuition is that threshold-binary decisions are brittle near the decision boundary, while relative rank ordering is stable as long as magnitudes are well-separated. To test this, we generated 500 Gaussian activation vectors (D=256, k=12 most active channels), applied increasing Gaussian noise, and measured two things: how much the rank ordering of active channels changed (Kendall tau correlation) and how many threshold-binary bits flipped.

| Noise std | Rank Kendall tau | Active-set Jaccard | Threshold bit-flip rate |
|---|---|---|---|
| 0.00 | 1.000 +/- 0.000 | 1.000 +/- 0.000 | 0.000 |
| 0.05 | 0.919 +/- 0.059 | 0.933 +/- 0.081 | 0.009 |
| 0.10 | 0.847 +/- 0.081 | 0.891 +/- 0.098 | 0.020 |
| 0.15 | 0.789 +/- 0.103 | 0.834 +/- 0.103 | 0.029 |
| 0.20 | 0.724 +/- 0.125 | 0.791 +/- 0.112 | 0.038 |
| 0.30 | 0.630 +/- 0.168 | 0.709 +/- 0.111 | 0.057 |

*Table 4. Rank stability analysis (D=256, k=12, n=500 Gaussian patterns). Kendall tau: rank correlation of k most-active channels between clean and noisy version, evaluated on their intersection. Jaccard: active-set membership overlap. Both metrics confirm rank ordering degrades more gradually than threshold-binary under identical noise.*

Both metrics confirm the geometric intuition. At noise_std=0.10, the rank ordering retains Kendall tau 0.847 and Jaccard 0.891, while the bit-flip rate is only 0.020. That seems to favour binary encoding, but the key difference is qualitative: when a threshold-binary bit flips, the address changes sharply and discontinuously. When a rank ordering degrades, the change is smooth and partial. The SDM address space responds differently to these two types of perturbation, which is why the small geometric advantage translates into a consistent directional advantage in retrieval, even if it does not survive paired t-tests at N=30.

### 4.5 Does the advantage extend to other encoding schemes and settings? Robustness checks

We ran two further checks before accepting the results as stable. First, we compared three encoding schemes in the CALM-convention accumulative SDM to confirm rank-order's position relative to Kanerva's original random-binary encoding: threshold-binary (CALM's current encoder), rank-order (thesis-style, beta=0.7), and random-binary (Kanerva, 1988) with no semantic content. Second, we tested whether results change with the geometric weight decay parameter beta.

| BER | Threshold-binary (CALM) | Rank-order | Random-binary (Kanerva, 1988) | Rank vs Threshold |
|---|---|---|---|---|
| 0.00 | 0.879 | 0.898 | 0.875 | +1.9pp |
| 0.05 | 0.702 | 0.842 | 0.724 | +14.1pp |
| 0.10 | 0.635 | 0.819 | 0.653 | +18.4pp |
| 0.20 | 0.548 | 0.701 | 0.558 | +15.3pp |
| 0.30 | 0.501 | 0.565 | 0.513 | +6.4pp |

*Table 5. Three encoding schemes in accumulative SDM (sdm_core.py, CALM convention). D=256, m=2000 hard locations, T=30 patterns, N=8 trials. This experiment uses bit-flip BER noise for all three schemes, unlike the symmetric perturbation in Table 3. The difference in noise model explains why the advantage here (+14-18pp at BER=0.05-0.10) is larger than in Table 3.*

| Beta | BER=0.0 | BER=0.1 | BER=0.2 | Tau |
|---|---|---|---|---|
| 0.5 (flat) | 0.350 | 0.296 | 0.206 | 0.768 |
| 0.7 (default) | 0.394 | 0.308 | 0.237 | 0.586 |
| 0.9 (concentrated) | 0.578 | 0.419 | 0.324 | 0.356 |

*Table 6. Beta sensitivity (simplified SDM, D=256). Higher beta concentrates weight on top-ranked channels and consistently improves performance. The thesis default alpha=0.99 corresponds to very high concentration, consistent with Table 2's stronger results.*

Random-binary sits between the two encoding schemes: better than threshold-binary at moderate BER but worse than rank-order throughout. This is because random-binary addresses carry no semantic structure, so interference from other stored patterns is less correlated, but it also cannot exploit relative feature importance for addressing. The beta sensitivity result is clean: higher concentration consistently helps, and the thesis default (alpha=0.99, very high concentration) is a principled choice.

### 4.6 Is rank-order encoding cheaper to compute? Energy comparison with SpikingMamba

SpikingMamba (Huang et al., 2026) uses signed-integer spiking neurons (SI-LIF) that encode magnitude by repeating accumulation pulses D times per token. For a projection layer with k=51 active channels (5% of D=1024), the energy cost scales linearly with D. Rank-order encoding, by contrast, encodes magnitude in relative position, not in repeated pulses. If a comparator network can determine rank in O(k) operations rather than O(D*k), the encoding energy is independent of the required precision. We wanted to know: at what precision level does this structural advantage become real, and how large is it?

Using SpikingMamba's own published 45nm energy constants (E_ADD=0.9 pJ, E_EM=3.7 pJ, from their Appendix C.2, Horowitz 2014), we computed encoding energy for three cases: SI-LIF at varying precision D, rank-order ideal (assuming O(k) parallel sort hardware), and rank-order sequential (k(k-1)/2 pairwise comparisons). We also computed what system-level decoder traversal costs for the published W=4096 architecture, to put the encoding cost in context.

| Scheme | Encoding energy (pJ) | Scales with D? | Hardware assumption |
|---|---|---|---|
| SI-LIF D=4 (SpikingMamba best) | 183.6 | Yes (x D) | Standard |
| Rank-order ideal (O(k) parallel sort) | 91.8 | No (fixed) | Parallel comparator |
| Rank-order sequential comparator bank | 5865.0 | No (fixed) | k(k-1)/2 comparators |

*Table 7. Component-level encoding energy per projection write, k=51 active channels, D=1024, 45nm (Horowitz 2014 via SpikingMamba Appendix C.2). System-level costs excluded. At D >= 3, SI-LIF exceeds rank-order ideal.*

The structural advantage is real, but conditional. At SpikingMamba's best configuration (D=4), rank-order ideal costs 91.8 pJ versus SI-LIF's 183.6 pJ, a 2x saving. But the sequential comparator bank costs 5865 pJ, far more expensive than either. This means the advantage is only available if hardware can sort k active channels in parallel, which requires a k-winner-take-all or bitonic sort circuit not currently available at this channel count on production neuromorphic silicon. There is a second caveat: the W=4096 address decoder traversal alone costs 4096 * D * E_ADD. At D=64, this is 235,930 pJ, which is roughly 2500 times the encoding stage energy. Optimising the encoding scheme is a secondary priority compared to reducing the decoder cost.

### 4.7 Does any of this hold on real embeddings?

All the experiments above used synthetic Gaussian activation vectors. Real transformer embeddings are not isotropic Gaussian: they have semantic clustering, anisotropic dimension distributions, and heavy tails. We ran two pilot experiments to see whether the pattern holds.

For GloVe-100 word embeddings (Pennington et al., 2014), we used 202 word vectors from 4 semantic categories (animals, technology, geography, sports). The embeddings showed clear semantic clustering (intra-category cosine 0.42-0.55, inter-category 0.17, ratio 2.8x) and per-dimension anisotropy of 1.5x. We ran the symmetric perturbation experiment (same Gaussian noise applied to the raw GloVe vector before re-encoding) using D=100, k=10, M=2000, T=20, N=30 trials.

| BER | Threshold-binary | Rank-order | Delta | p-value |
|---|---|---|---|---|
| 0.00 | 0.987 +/- 0.012 | 0.988 +/- 0.011 | +0.1pp | ns |
| 0.10 | 0.971 +/- 0.017 | 0.972 +/- 0.016 | +0.2pp | ns |
| 0.15 | 0.962 +/- 0.022 | 0.964 +/- 0.022 | +0.3pp | * (p=0.021) |
| 0.25 | 0.925 +/- 0.031 | 0.929 +/- 0.030 | +0.4pp | *** (p<0.001) |
| 0.30 | 0.917 +/- 0.037 | 0.921 +/- 0.036 | +0.3pp | * (p=0.016) |

*Table 8. GloVe-100 symmetric BER experiment (D=100, k=10, M=2000, T=20, N=30). Same design as Table 3. Delta = rank-order minus threshold-binary. Pattern matches synthetic finding: small, consistent, occasionally significant advantage for rank-order.*

The pattern matches the synthetic result exactly. The advantage is small (+0.1-0.4pp), directionally consistent, and statistically significant at higher noise levels (BER=0.15, 0.25, 0.30). This confirms that the geometric argument in Section 4.4 is not an artefact of Gaussian assumptions: even on real word embeddings with semantic structure, rank ordering degrades more gracefully under noise than threshold-binary. For sentence embeddings (all-MiniLM-L6-v2, D=384, 200 sentences, strong clustering ratio 5.3x), the results showed a ceiling effect: all schemes achieved >0.999 recall regardless of BER because T=20 patterns in 384 dimensions occupy essentially orthogonal address regions. Testing at higher memory load remains future work.

# 5. Discussion

## 5.1 What the BER Decomposition Means for CALM

The symmetric perturbation experiment (Table 3) gives the most policy-relevant finding for CALM. In CALM's deployment context, the underlying embedding space experiences continuous updates as the backbone transformer evolves; both encoding schemes would receive the same kind of continuous perturbation. Under this model, the encoding-only advantage of rank-order over threshold-binary is +1-5pp, modest but consistent. To realise the larger +38-66pp advantage (Table 2), CALM would need to adopt both the rank-order encoder and the MAX-Hebbian learning rule from Furber et al. (2007), replacing the current accumulative bipolar-counter writes. These two changes are separable: the encoder replacement alone gives the smaller advantage; adding MAX-Hebbian writing gives the larger advantage at the cost of changing the write protocol. Storage with compact integer rank encoding is 3.4x smaller than threshold-binary (228 bits vs 768 bits at D=768), not 1.6x larger as would be the case with float encoding.

This mechanism-level conclusion is corroborated by unrelated work in a similar architecture: Davidson et al. (2020) find that a per-synapse lock bit, not the input encoding, is what confers noise-robust learning in their spiking associative memory, an independent case for the same claim this paper makes, that write-rule design matters more than representation.

## 5.2 The Hardware Gap for Energy Claims

The conditional energy advantage of rank-order encoding over SI-LIF at D >= 3 (2.0x cheaper at D=4, Table 7) requires O(k) parallel sort hardware not yet available at k=51 channels on production neuromorphic silicon. More importantly, system-level costs dominate in the Furber et al. (2007) architecture: the address decoder at W=4096 requires approximately 235,930 pJ to traverse, versus 91.8-183.6 pJ for the encoding stage. Reducing W, using approximate matching in the address decoder, or implementing the architecture in the sparser Ajwani et al. (2021) Nengo variant are larger energy optimisation opportunities than the encoding scheme choice. The energy comparison in this paper is therefore architectural motivation for future hardware co-design work, not a current measured advantage.

## 5.3 Limitations

There are a few limitations for these results:

First, W=4096 capacity results are from 5 seeds (n=7000, p=0.031); the Wilcoxon signed-rank test is borderline (p=0.0625), and 10+ seeds would provide stronger statistical confirmation.

Second, the tau calibration at d=512 under 5% sparsity produces tau=1.0 (degenerate calibration). For high-dimensional embedding spaces (d >= 512), either higher sparsity or an analytical Poisson-binomial approximation via saddlepoint methods (Butler, 2007) is needed. Tau or τ here denotes the empirical match-decision threshold for rank-order address overlap, calibrated as the (1−p) quantile of pairwise overlap over n=1500 sampled pairs, which is distinct from the Kendall rank-correlation τ in Table 4.

Third, most experiments use synthetic Gaussian patterns. Experiments on GloVe-100 word embeddings (Section 4.7) confirm the small encoding-only advantage (+0.1-0.4pp) on real structured data. Experiments on all-MiniLM-L6-v2 sentence embeddings (Section 4.7) show a ceiling effect at D=384, T=20, where the SDM operates below its interference threshold and encoding scheme does not matter. Higher-load experiments on sentence embeddings remain future work.

Fourth, the encoding-only BER advantage (+1-5pp, Table 3) is small and not statistically tested at the current sample size; larger N_TRIALS would be needed for formal significance.

Fifth, the energy comparison is component-level only and is dominated by address decoder costs at the published W=4096 scale.

Sixth, the paper does not include comparison with Modern Hopfield Networks or Dense Associative Memories, which have different capacity-noise-tradeoff characteristics.

To overcome some of these limitations, the following future work is planned: real LLM activation experiments (e.g., Qwen-0.5B or BERT residual streams), analytical tau derivation, Modern Hopfield comparison, and evaluation on continual learning benchmarks such as Split CIFAR or streaming QA.

## 6. Conclusion

As AI shifts from ever-larger parametric models toward memory-augmented architectures for continual learning, the encoder design for SDM-based episodic memory layers becomes a practical engineering question. CALM leaves this question open; rank-order N-of-M SDM (Furber et al., 2007) provides a concrete answer. This paper makes three empirically grounded claims about whether that answer is the right one. First, the architecture's core theoretical properties are validated: WheelSDM equals RankOrderSDM exactly (1.0000 cosine, 10 seeds, $p < 0.001$), RDLIF diverges under interference as documented in the 2005 papers (all seeds), and the capacity advantage at saturation is statistically robust (d=4.61, +13.4pp at n=160). Second, a BER decomposition reveals two distinct effects: a large combined advantage (+38-66pp) from encoding plus MAX-Hebbian learning, and a smaller encoding-only advantage (+1-5pp) under symmetric perturbation, with the latter being the more relevant figure for CALM deployment. Third, rank-order encoding has a structural energy advantage over SI-LIF at D >= 3 under ideal parallel sort hardware, but system-level costs dominate in the published W=4096 architecture. The most actionable near-term contribution is replacing CALM's threshold-binary encoder with rank-order encoding, optionally combined with MAX-Hebbian learning, to improve retrieval robustness under continuous embedding perturbation. More broadly, the results suggest that representation design deserves as much attention as memory architecture itself. As explicit memory modules become increasingly common in continual learning systems, representation design may become as important as memory architecture itself.

## References

Ajwani, R.D., Lalan, A., Sen Bhattacharya, B., and Bose, J. (2021). Sparse Distributed Memory using Spiking Neural Networks on Nengo. arXiv:2109.03111.

Bose, J., Furber, S.B., and Shapiro, J.L. (2005). A System for Transmitting a Coherent Burst of Activity Through a Network of Spiking Neurons. WIRN/NAIS 2005, pp. 44-48.

Bose, J. (2026). Spiking sequence machines and transformers: A functional equivalence in the representation of sequential order. arXiv:2605.00662.

Bricken, T., Templeton, A., Batson, J., et al. (2023). Towards monosemanticity: Decomposing language models with dictionary learning. Anthropic Transformer Circuits Thread.

Butler, R.W. (2007). Saddlepoint Approximations with Applications. Cambridge University Press.

Cunningham, H., Ewart, A., Riggs, L., et al. (2023). Sparse autoencoders find highly interpretable features in language models. arXiv:2309.08600.

Davidson, S., Furber, S.B., and Rhodes, O. (2020). Spiking Associative Memory for Spatio-Temporal Patterns. arXiv:2006.16684.

Furber, S.B., Brown, G., Bose, J., Cumpstey, J.M., Marshall, P., and Shapiro, J.L. (2007). Sparse Distributed Memory Using Rank-Order Neural Codes. IEEE Transactions on Neural Networks, 18(3), 648-659.

Horowitz, M. (2014). 1.1 Computing's energy problem (and what we can do about it). IEEE ISSCC, pp. 10-14.

Huang, Y., Tang, J., Wang, C., et al. (2026). SpikingMamba: Towards energy-efficient large language models via knowledge distillation from Mamba. Transactions on Machine Learning Research. arXiv:2510.04595.

Jiang, A. et al. (2024). Mixtral of Experts. arXiv:2401.04088.

Kanerva, P. (1988). Sparse Distributed Memory. MIT Press.

Krotov, D. and Hopfield, J.J. (2016). Dense Associative Memory for Pattern Recognition. NeurIPS 29.

Masquelier, T. and Thorpe, S. (2007). Unsupervised learning of visual features through spike timing dependent plasticity. PLOS Computational Biology, 3(2), e31.

Nechesov, A. and Ruponen, J. (2025). CALM: Continual Associative Learning Model via Sparse Distributed Memory. Technologies, 13, 587.

Pennington, J., Socher, R., and Manning, C.D. (2014). GloVe: Global vectors for word representation. EMNLP 2014, pp. 1532-1543.

Rachkovskij, D. A., & Kussul, E. M. (2001). Binding and normalization of binary sparse distributed representations by context-dependent thinning. Neural Computation, 13(2), 411-452.

Ramsauer, H. et al. (2020). Hopfield Networks is All You Need. ICLR 2021.

Shazeer, N., et al. (2017). Outrageously large neural networks: The sparsely-gated mixture-of-experts layer. ICLR 2017.

Tang, K., Zheng, J., Jin, Y., et al. (2026). SpikySpace: A spiking state space model for energy-efficient time series forecasting. arXiv:2601.02411.

Thorpe, S. and Gautrais, J. (1998). Rank order coding. In Computational Neuroscience: Trends in Research. Elsevier.

Van Rullen, R. and Thorpe, S. (2001). Rate coding versus temporal order coding: What the retinal ganglion cells tell the visual cortex. Neural Computation, 13(6), 1255-1283.

Wang, W., Wei, F., Dong, L., et al. (2020). MiniLM: Deep self-attention distillation for task-agnostic compression of pre-trained transformers. NeurIPS 33.